\documentclass[fleqn,10pt]{wlscirep}
\usepackage[utf8]{inputenc}
\usepackage[T1]{fontenc}
\usepackage{amsmath}
\usepackage{bm}
\usepackage{multirow}
\usepackage{colortbl}
\usepackage{microtype}
\usepackage{pgfplots, pgfplotstable}
\usepackage{mwe}
\usepackage{float}
\usepackage{lineno}
\usepackage{nicefrac}
\usepackage{color}
\usepackage{transparent}
\graphicspath{{pdf/}}
\usepackage{siunitx}
\usepackage{subcaption}
\pgfplotsset{compat=1.15}

\usepackage[nolist,nohyperlinks]{acronym}

\begin{acronym}
    \acro{DL}[DL]{deep learning}
    \acro{NIH}[NIH]{National Institutes of Health}
    \acro{SNN}[SNN]{siamese neural network}
    \acro{FC}[FC]{fully-connected}
    \acro{BCE}[BCE]{binary cross-entropy}
    \acro{CE}[CE]{cross-entropy}
    \acro{Adam}[Adam]{adaptive moment estimation}
    \acro{SGD}[SGD]{stochastic gradient descent}
    \acro{LR}[LR]{learning rate}
    \acro{ROC}[ROC]{receiver operating characteristic}
    \acro{AUC}[AUC]{area under the curve}
    \acro{TPR}[TPR]{true positive rate}
    \acro{FPR}[FPR]{false positive rate}
    \acro{TP}[TP]{true positive}
    \acro{FP}[FP]{false positive}
    \acro{FN}[FN]{false negative}
    \acro{TN}[TN]{true negative}
    \acro{DP}[DP]{differential privacy}
    \acro{FL}[FL]{federated learning}
    \acro{ASV}[ASV]{automatic speaker verification}
\end{acronym}

\pgfplotstableread[col sep=comma,]{./csv/Siamese_ResNet50_Set_full_LR_-4.csv}\roctableSizeEightHundred
\pgfplotstableread[col sep=comma,]{./csv/Siamese_ResNet50_Set_400k_LR_-4.csv}\roctableSizeFourHundred
\pgfplotstableread[col sep=comma,]{./csv/Siamese_ResNet50_Set_200k_LR_-4.csv}\roctableSizeTwoHundred
\pgfplotstableread[col sep=comma,]{./csv/Siamese_ResNet50_Set_100k_LR_-4.csv}\roctableSizeOneHundred
\pgfplotstableread[col sep=comma,]{./csv/Siamese_ResNet50_Set_full_LR_-7.csv}\roctableLRSeven
\pgfplotstableread[col sep=comma,]{./csv/Siamese_ResNet50_Set_full_LR_-6.csv}\roctableLRSix
\pgfplotstableread[col sep=comma,]{./csv/Siamese_ResNet50_Set_full_LR_-5.csv}\roctableLRFive
\pgfplotstableread[col sep=comma,]{./csv/Siamese_ResNet50_Set_full_LR_-3.csv}\roctableLRThree

\newlength{\tempheight}
\newlength{\tempwidth}

\newcommand{\rowname}[1]% #1 = text
{\rotatebox{90}{\makebox[\tempheight][c]{\textbf{#1}}}}

\newcommand{\columnname}[1]% #1 = text
{\makebox[\tempwidth][c]{\textbf{#1}}}

\usepackage{hyperref}
\usepackage[capitalise,noabbrev]{cleveref}

\title{Deep Learning-based Patient Re-identification Is able to Exploit the Biometric Nature of Medical Chest X-ray Data}

\author[1,*]{Kai Packhäuser}
\author[1]{Sebastian Gündel}
\author[1]{Nicolas Münster}
\author[1]{Christopher Syben}
\author[1]{Vincent Christlein}
\author[1]{Andreas Maier}
\affil[1]{Pattern Recognition Lab, Department of Computer Science, Friedrich-Alexander-Universität Erlangen-Nürnberg, 91058 Erlangen, Germany}

\affil[*]{kai.packhaeuser@fau.de}

\begin{abstract}
\bfseries
With the rise and ever-increasing potential of deep learning techniques in recent years, publicly available medical datasets became a key factor to enable reproducible development of diagnostic algorithms in the medical domain. Medical data contains sensitive patient-related information and is therefore usually anonymized by removing patient identifiers, e.\,g., patient names before publication. To the best of our knowledge, we are the first to show that a well-trained deep learning system is able to recover the patient identity from chest X-ray data. We demonstrate this using the publicly available large-scale ChestX-ray14 dataset, a collection of 112,120 frontal-view chest X-ray images from 30,805 unique patients. Our verification system is able to identify whether two frontal chest X-ray images are from the same person with an AUC of 0.9940 and a classification accuracy of 95.55\,\%. We further highlight that the proposed system is able to reveal the same person even ten and more years after the initial scan. When pursuing a retrieval approach, we observe an mAP@R of 0.9748 and a precision@1 of 0.9963. Furthermore, we achieve an AUC of up to 0.9870 and a precision@1 of up to 0.9444 when evaluating our trained networks on external datasets such as CheXpert and the COVID-19 Image Data Collection. Based on this high identification rate, a potential attacker may leak patient-related information and additionally cross-reference images to obtain more information. Thus, there is a great risk of sensitive content falling into unauthorized hands or being disseminated against the will of the concerned patients. Especially during the COVID-19 pandemic, numerous chest X-ray datasets have been published to advance research. Therefore, such data may be vulnerable to potential attacks by deep learning-based re-identification algorithms.\\[-0.4cm]
\end{abstract}

\begin{document}

\flushbottom
\maketitle

%
% INTRODUCTION
%
\noindent
Chest radiography (X-ray) is a modality that is routinely used for diagnostic procedures around the world~\cite{maier2018medical}.
It became the most common medical imaging examination for pulmonary diseases and allows a clear investigation of the thorax~\cite{raoof2012interpretation}. Chest X-ray imaging is therefore well-suited for diagnosing several pathologies including pulmonary nodules, masses, pleural effusions, pneumonia, COPD, and cardiac abnormalities~\cite{guendel2018learning}. It is also used for COVID-19~\cite{WHOcorona} screening, as abnormalities typical of those infected with the coronavirus can be detected in radiographs~\cite{wang2020covid}. While chest radiography plays a crucial role in clinical care, discovering certain diseases and abnormalities in chest radiographs can be a challenging task for radiologists, which potentially results in undesirable misdiagnoses~\cite{lee2013cognitive}. Therefore, computer-aided detection~(CAD) systems based on \ac{DL}~\cite{lecun2015deep} techniques have been developed in recent years to facilitate radiology workflows. These systems, characterized by their enormous benefits, can be utilized for a wide range of applications, e.\,g., for the automatic recognition of abnormalities in chest radiographs~\cite{guendel2018learning,guendel2019multi} and the detection of tumors in mammography~\cite{akselrod2016region}. Some techniques even show the potential to exceed human performance~\cite{rajpurkar2017chexnet}. However, the CAD systems are only treated as an additional source to support the radiologists and to increase certainty in their reading decisions.
\newline
\indent
On the one hand, the large variety of medical applications allows \ac{DL} to grow and tackle real-life problems that were previously not solvable or improving solutions offered by traditional machine learning methods~\cite{lecun2015deep}. On the other hand, \ac{DL} is a data-driven approach and well-known for its need for big data to train the neural networks~\cite{roh2019survey, maier2019gentle}. For these reasons, a vast amount of medical datasets have been published in recent years that enable researchers to develop diagnostic algorithms in the medical field in a reproducible way~\cite{oakden2020exploring}. These include several large-scale chest radiography datasets, e.\,g., the \textit{CheXpert}~\cite{irvin2019chexpert}, the \textit{PLCO}~\cite{team2000prostate} and the \textit{ChestX-ray14}~\cite{wang2017chestx} datasets. But especially during the COVID-19 pandemic~\cite{bandyopadhyay2020covid,spinelli2020covid}, the number of publicly available chest radiography datasets increased rapidly. A few selected examples are the \textit{COVID-19 Image Data Collection}~\cite{cohen2020covid}, the \textit{Figure 1 COVID-19 Chest X-ray Dataset Initiative}~\cite{figure1covid}, the \textit{ActualMed COVID-19 Chest X-ray Dataset Initiative}~\cite{ActualMedCovid}, and the \textit{COVID-19 Radiography~Database}~\cite{RadiographyDatabase}. 
\newline
\indent
Chest radiography datasets typically consist of two parts: First, the image data itself, which provides clinical information about the anatomical structure of the thorax. Second, the associated metadata, which contains sensitive patient-related information that is either stored in a separate file or embedded directly in the images~\cite{willemink2020preparing}.
Proper data anonymization constitutes an important step when preparing medical data for public usage to ensure that a patient's identity cannot be revealed in publicly available datasets~\cite{willemink2020preparing}. 
In practice, any personally identifiable information is attempted to be removed from the data before it is shared.
These objectives and requirements are specified, e.\,g., by the Health Insurance Portability and Accountability Act (HIPAA)~\cite{hipaa} in the United States or the General Data Protection Regulation (GDPR)~\cite{gdpr} in Europe.
\newline
\indent
In 2017, Google entered into a project with the \ac{NIH} to publish a dataset containing 100,000 chest radiographs. However, the release was canceled two days before publication after Google was informed by the \ac{NIH} that the radiographs still contained personal information which indicates that the data was incorrectly anonymized~\cite{healthimaging,theverge}. This major incident highlights that many potential pitfalls can arise when clinical and technological institutions collect and share large medical datasets to revolutionize health-care.
\newline
\indent
In the past, various data de-identification techniques have been proposed, including commonly-used methods like pseudonymization~\cite{noumeir2007pseudonymization} and $k$-anonymity~\cite{sweeney2002k}. Pseudonymization describes a technique that replaces a true identifier, e.\,g., the name or the patient identification number by a pseudonym that is unique to the patient but has no relation to the person~\cite{noumeir2007pseudonymization}. However, pseudonymization is a rather weak anonymization technique as the patient's identity may still be revealed, e.\,g., by cross-referencing with other publicly available datasets. In contrast, $k$-anonymity modifies the data before sharing in such a way that every sample in the published dataset can be associated with at least $k$ different subjects. In this way, the probability of performing identity disclosure is limited to at most $\nicefrac{1}{k}$~\cite{sweeney2002k,gkoulalas2015medical}. Nevertheless, when background knowledge is available, $k$-anonymity is susceptible to many attacks.
\begin{figure}
    \centering
    \includegraphics[width=0.85\textwidth]{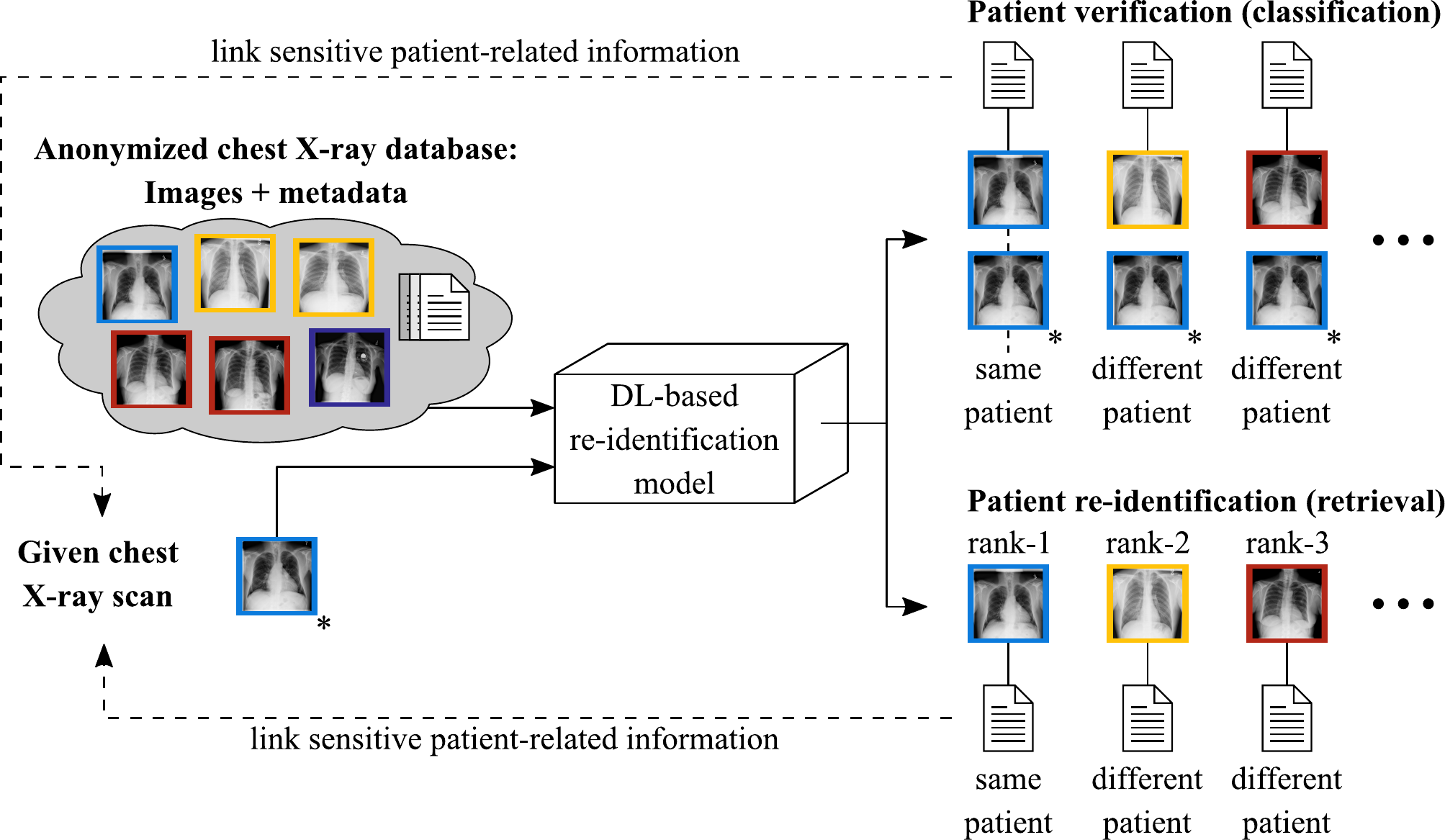}
    \caption{General problem scenario: Comparing a given chest radiograph to publicly available dataset images by means of \ac{DL} techniques would either result in discrete labels indicating whether or not the dataset images belong to the same patient as the given radiograph (verification scenario) or yield a ranked list of the most similar radiographs related to the given scan (retrieval scenario). Images belonging to the same patient are highlighted with the same color. The given radiograph is marked with an asterisk. The shown cases would enable a potential attacker to link sensitive patient-related information contained in the dataset to the image of interest.}
    \label{fig:overview_graphics}
\end{figure}
\newline
\indent
To date, little attention has been paid to the possibility of re-identifying patients in large medical datasets by means of \ac{DL} techniques. However in theory, medical data disclosure, as illustrated in \cref{fig:overview_graphics}, could be facilitated for potential attackers by using suitable \ac{DL} approaches. Consider a publicly available dataset that is supposedly anonymized but contains further sensitive patient-related information, e.\,g., diagnosis, treatment history, and clinical institution. If a radiograph of known identity is accessible to a potential attacker and a properly working verification or re-identification model exists, then the model could be used to compare the given radiograph to each image in the dataset which would essentially result in a set of images belonging to the same patient (patient verification) or yield a ranked list of the most similar images to the given radiograph (patient re-identification). In this way, the patient's identity may be linked to sensitive data contained in the dataset. As a result, more patient-related information may have been leaked, highlighting the enormous data security and data privacy issues involved.
\newline
\indent
In our work, we investigated whether conventional anonymization techniques are secure enough and whether it is possible to re-identify and de-anonymize individuals from their medical data using \ac{DL}-based methods. Therefore, we considered the public ChestX-ray14 dataset~\cite{wang2017chestx}, which is one of the most widely used research datasets in the field of radiographic problems. Our algorithms are trained to determine whether two arbitrary chest radiographs can be recognized to belong to the same patient or not. Moreover, we showed that our proposed methods are able to perform a successful linkage attack on publicly available chest radiography datasets. Furthermore, this work aims to draw attention to the massive problem of releasing medical data without considering that \ac{DL} systems can easily be used to reveal a patient's identity. Therefore, we call for reconsidering conventional anonymization techniques and developing more secure methods that resist potential attacks by \ac{DL} algorithms.

%
% RESULTS
%
\section*{Patient Verification}
    \begin{table}[tb]
        \centering
        \caption{Overview of the obtained verification results for our experiments using varying training set sizes $N_s$ at different learning rates $\eta$. Moreover, different data handling techniques were used (FTS: Fixed training set; RNP: Randomized negative pairs). For each experiment, the training sets were balanced with respect to the amount of positive and negative image pairs. In this table, we present the \ac{AUC} (together with the lower and upper bounds of the \SI{95}{\percent} confidence intervals from 10,000 bootstrap runs), the accuracy, the specificity, the recall, the precision, and the F1-score. Bold text emphasizes the overall highest \ac{AUC} value.}
        \label{tab:results}
        \resizebox{\textwidth}{!}{
        \renewcommand{\arraystretch}{1.3}
\begin{tabular}{|c|c|c|c|c|c|c|c|c|}
\hline
\rowcolor[HTML]{C0C0C0} 
\textbf{\begin{tabular}[c]{@{}c@{}}Data\\ handling\end{tabular}} & \textbf{N\textsubscript{s}} & \textbf{$\eta$} & \textbf{AUC + 95\,\% CI} & \textbf{Accuracy ($\frac{TP+TN}{P+N}$)} & \textbf{Specificity ($\frac{TN}{N}$)} & \textbf{Recall ($\frac{TP}{P}$)} & \textbf{Precision ($\frac{TP}{TP+FP}$)} & \textbf{F1-score} \\ \hline
                      & 100,000 & $10^{-3}$ & $0.8610\enspace_{0.8588}^{0.8632}$          & $0.7782\enspace (\frac{77,815}{100,000})$   & $0.7710\enspace (\frac{38,548}{50,000})$       & $0.7853\enspace (\frac{39,267}{50,000})$     & $0.7742\enspace (\frac{39,267}{50,719})$             & $0.7797$ \\ \cline{2-9} 
                      & 200,000 & $10^{-3}$ & $0.9448\enspace_{0.9435}^{0.9461}$          & $0.8743\enspace (\frac{87,428}{100,000})$   & $0.8685\enspace (\frac{43,426}{50,000})$       & $0.8800\enspace (\frac{44,002}{50,000})$     & $0.8700\enspace (\frac{44,002}{50,576})$             & $0.8750$ \\ \cline{2-9} 
                      & 400,000 & $10^{-4}$ & $0.9587\enspace_{0.9575}^{0.9599}$          & $0.8755\enspace (\frac{87,546}{100,000})$   & $0.9290\enspace (\frac{46,452}{50,000})$       & $0.8219\enspace (\frac{41,094}{50,000})$     & $0.9205\enspace (\frac{41,094}{44,642})$             & $0.8684$ \\ \cline{2-9} 
\multirow{-4}{*}{FTS} & 800,000 & $10^{-4}$ & $0.9896\enspace_{0.9891}^{0.9901}$          & $0.9537\enspace (\frac{95,367}{100,000})$   & $0.9541\enspace (\frac{47,705}{50,000})$       & $0.9532\enspace (\frac{47,662}{50,000})$     & $0.9541\enspace (\frac{47,662}{49,957})$             & $0.9536$ \\ \hline
RNP                   & 800,000 & $10^{-4}$ & $\textbf{0.9940}\enspace_{0.9937}^{0.9944}$ & $0.9555\enspace (\frac{95,545}{100,000})$   & $0.9822\enspace (\frac{49,111}{50,000})$       & $0.9287\enspace (\frac{46,434}{50,000})$     & $0.9812\enspace (\frac{46,434}{47,323})$             & $0.9542$ \\ \hline
\end{tabular}
        }
    \end{table}
First, we trained a \ac{SNN} architecture on the ChestX-ray14 dataset to determine whether two individual chest radiographs correspond to the same patient or not. Our model was designed to process the two input images in two identical network branches, which are then combined by a merging layer. The fused information is fed through further network layers resulting in a single output score indicating the identity similarity.
\newline
\indent
\Cref{tab:results} summarizes the outcomes of our evaluation. We analyzed a multitude of different experimental setups with varying \acl{LR}s~$\eta$ and differing balanced training set sizes~$N_s$. Moreover, we investigated the effect of using epoch-wise randomized negative pairs (RNP) versus fixed training sets (FTS) for the entire learning procedure. When using RNP as the data handling technique, the negative image pairs were randomly constructed in each epoch, meaning that much more negative pairs could be utilized in a complete training run compared to FTS where the generated image pairs remain the same for the entire learning procedure. For all experiments on the ChestX-ray14 dataset, we used the same balanced validation and testing set with 50,000 and 100,000 image pairs, respectively, without patient overlap between any split. To assess the performance of the trained models, we performed a \ac{ROC} analysis by computing the \ac{AUC} value together with the \SI{95}{\percent} confidence intervals from 10,000 bootstrap runs. Moreover, we calculated the accuracy, specificity, recall, precision, and F1-score.
    \begin{figure}[tb]
        %\centering
        \hfill
        \begin{minipage}[b]{.45\textwidth}
            \centering
            \resizebox{\textwidth}{!}{
            \begin{tikzpicture}
\pgfplotsset{%
    width=\textwidth
}
\begin{axis}[
  grid=both,
  xlabel={False Positive Rate},
  ylabel={True Positive Rate},
  enlargelimits=false,
  clip=false,
  legend pos=south east
]
\addplot[thick,color=black!50!green] table [x=X, y=Y]{\roctableSizeEightHundred};
\addplot[thick,color=blue] table [x=X, y=Y]{\roctableSizeFourHundred};
\addplot[thick,color=red] table [x=X, y=Y]{\roctableSizeTwoHundred};
\addplot[thick,color=orange] table [x=X, y=Y]{\roctableSizeOneHundred};
\addlegendentry{$N=800,000$}
\addlegendentry{$N=400,000$}
\addlegendentry{$N=200,000$}
\addlegendentry{$N=100,000$}
\end{axis}
\end{tikzpicture}
            }
            \captionof{figure}{ROC curves for different training set sizes $N_s$ and a fixed LR of $\eta=10^{-4}$. During training, the fixed data handling technique was employed.}
            \label{fig:ROC_curves_training_size}
        \end{minipage}%
        \hfill
        \begin{minipage}[b]{.45\textwidth}
            \centering
            \resizebox{\textwidth}{!}{
            \renewcommand{\arraystretch}{2.0}
\begin{tabular}{|l|c|c|c|c}
\cline{3-4}
\multicolumn{2}{c|}{}&\multicolumn{2}{c|}{\textbf{Predicted}}&\\
\cline{3-5}
\multicolumn{2}{c|}{}&neg&pos&\multicolumn{1}{c|}{\textbf{Total}}\\
\cline{1-5}
\multirow{2}{*}{\rotatebox[origin=c]{90}{\textbf{Actual}}}& neg & $49,111$ (TN) & $\color{white}44,\color{black}889$ (FP) & \multicolumn{1}{c|}{$50,000$}\\
\cline{2-5}
& pos & $\color{white}4\color{black}3,566 $ (FN) & $46,434$ (TP) & \multicolumn{1}{c|}{$50,000$}\\
\cline{1-5}
\multicolumn{1}{c|}{} & \multicolumn{1}{c|}{\textbf{Total}} & \multicolumn{1}{c|}{$52,677$} & \multicolumn{    1}{c|}{$47,323$} & \multicolumn{1}{c|}{$100,000$}\\
\cline{2-5}
\end{tabular}
            }
            \vspace{0.9cm}
            \captionof{figure}{Confusion matrix corresponding to the best experiment shown in \cref{tab:results} (last row) giving clear insights into the performance of our trained model.}
            \label{fig:confusion_matrix}
        \end{minipage}
        \hfill
    \end{figure}
\newline
\indent
The results indicate that the amount of training data plays a crucial role in the patient verification task. We observe a significant performance increase as the training set size grows. For instance, when using a subset of 100,000 image pairs for training, we obtain an \ac{AUC} value of 0.8610. In contrast, by enlarging the training set size to 800,000 image pairs (i.\,e.\ the total number of 400,000 positive image pairs combined with 400,000 negative pairs), we receive an \ac{AUC} score of 0.9896. These findings have been visualized in the \ac{ROC} curves shown in \cref{fig:ROC_curves_training_size} which illustrates the effect of the training set size on the verification performance when using fixed training sets. Note that \cref{tab:results} only shows the best experiments per training set size~$N_s$. Additional experiments were conducted to investigate the effect of the \ac{LR}. The corresponding results are provided in a separate table in the appendix (see Supplementary Table~1).
%\ref{tab:additional_results}).
\newline
\indent
We also observed that randomly constructing the negative image pairs in each epoch led to further improvements in the final model performance. By using this data handling technique, we achieved our overall best results. The respective outcomes are reported in \cref{tab:results}. When training our network architecture with a total of 800,000 training samples with epoch-wise randomly constructed negative pairs, the \ac{AUC} score improved from 0.9896 to 0.9940. Besides, the other reported evaluation metrics apart from the recall also increased compared to the results achieved by the model trained with the fixed set. \Cref{fig:confusion_matrix} depicts the confusion matrix resulting from our best-trained model listed in \cref{tab:results} (last row), thus giving clear insights into the patient verification performance.
    \begin{figure}[t!]
        \centering
        \resizebox{\textwidth}{!}{
\begin{tikzpicture}[scale=.7]

  \draw (0cm,0cm) -- (16.5cm,0cm);  %Abzisse
  \draw (0cm,0cm) -- (0cm,-0.1cm);
  \draw (16.5cm,0cm) -- (16.5cm,-0.1cm);
  
  \draw (-0.1cm,0cm) -- (-0.1cm,6cm);  %Ordinate
  \draw (-0.1cm,0cm) -- (-0.2cm,0cm); 
  \draw (-0.1cm,6cm) -- (-0.2cm,6cm); %node [left] {\%};
  
  \draw (18.0cm,0cm) -- (20.75cm,0cm);  %Abzisse
  \draw (18.0cm,0cm) -- (18.0cm,-0.1cm);
  \draw (20.75cm,0cm) -- (20.75cm,-0.1cm);
  
  \draw (22.25cm,0cm) -- (25.0cm,0cm);  %Abzisse
  \draw (22.25cm,0cm) -- (22.25cm,-0.1cm);
  \draw (25.0cm,0cm) -- (25.0cm,-0.1cm);
  
  \draw[gray!50, text=black] (-0.2 cm,1 cm) -- (25 cm,1 cm) 
      node at (-1.0 cm,1 cm) {0.2};  %Beschriftung der Hilfslinien
  \draw[gray!50, text=black] (-0.2 cm,2 cm) -- (25 cm,2 cm) 
      node at (-1.0 cm,2 cm) {0.4};  %Beschriftung der Hilfslinien
  \draw[gray!50, text=black] (-0.2 cm,3 cm) -- (25 cm,3 cm) 
      node at (-1.0 cm,3 cm) {0.6};  %Beschriftung der Hilfslinien
  \draw[gray!50, text=black] (-0.2 cm,4 cm) -- (25 cm,4 cm) 
      node at (-1.0 cm,4 cm) {0.8};  %Beschriftung der Hilfslinien
  \draw[gray!50, text=black] (-0.2 cm,5 cm) -- (25 cm,5 cm) 
      node at (-1.0 cm,5 cm) {1.0};  %Beschriftung der Hilfslinien
      
  \draw[fill=blue!40] (0.5 cm,0cm) rectangle (0.5cm+0.5 cm,4.84375 cm) %die Säulen
    node at (0.5cm + 0.25 cm,4.84375 cm + 0.3cm) {\small{0.97}}; %die Prozente über den Säulen 
  \node[] at (0.75 cm,-0.5cm) {< 1};%Säulenbeschriftung
  \node[rotate=90] at (0.75 cm,1.5cm) {\footnotesize{31317 / 32327}};
  
  \draw[fill=blue!40] (1.75 cm,0cm) rectangle (1.75cm+0.5 cm,4.70065 cm) %die Säulen
    node at (1.75cm + 0.25 cm,4.70065 cm + 0.3cm) {\small{0.94}}; %die Prozente über den Säulen
  \node[] at (1.75 cm + 0.25cm,-0.5cm) {1}; %Säulenbeschriftung
  \node[rotate=90] at (1.75 cm + 0.25cm,1.5cm) {\footnotesize{4774 / 5078}};
  
  \draw[fill=blue!40] (3.0 cm,0cm) rectangle (3.0cm+0.5 cm,4.71715 cm) %die Säulen
    node at (3.0cm + 0.25 cm,4.71715 cm + 0.3cm) {\small{0.94}}; %die Prozente über den Säulen
  \node[] at (3.0 cm + 0.25cm,-0.5cm) {2}; %Säulenbeschriftung
  \node[rotate=90] at (3.0 cm + 0.25cm,1.5cm) {\footnotesize{3019 / 3200}};
  
  \draw[fill=blue!40] (4.25 cm,0cm) rectangle (4.25cm+0.5 cm,4.77005 cm) %die Säulen
    node at (4.25cm + 0.25 cm,4.77005 cm + 0.3cm) {\small{0.95}}; %die Prozente über den Säulen
  \node[] at (4.25 cm + 0.25cm,-0.5cm) {3}; %Säulenbeschriftung
  \node[rotate=90] at (4.25 cm + 0.25cm,1.5cm) {\footnotesize{1722 / 1805}};
  
  \draw[fill=blue!40] (5.5 cm,0cm) rectangle (5.5cm+0.5 cm,4.54815 cm) %die Säulen
    node at (5.5cm + 0.25 cm,4.54815 cm + 0.3cm) {\small{0.91}}; %die Prozente über den Säulen
  \node[] at (5.5 cm + 0.25cm,-0.5cm) {4}; %Säulenbeschriftung
  \node[rotate=90] at (5.5 cm + 0.25cm,1.5cm) {\footnotesize{2426 / 2667}};
  
  \draw[fill=blue!40] (6.75 cm,0cm) rectangle (6.75cm+0.5 cm,4.4926 cm) %die Säulen
    node at (6.75cm + 0.25 cm,4.4926 cm + 0.3cm) {\small{0.90}}; %die Prozente über den Säulen
  \node[] at (6.75 cm + 0.25cm,-0.5cm) {5}; %Säulenbeschriftung
  \node[rotate=90] at (6.75 cm + 0.25cm,1.5cm) {\footnotesize{1700 / 1892}};
  
  \draw[fill=blue!40] (8.0 cm,0cm) rectangle (8.0cm+0.5 cm,4.6584 cm) %die Säulen
    node at (8.0cm + 0.25 cm,4.6584 cm + 0.3cm) {\small{0.93}}; %die Prozente über den Säulen
  \node[] at (8.0 cm + 0.25cm,-0.5cm) {6}; %Säulenbeschriftung
  \node[rotate=90] at (8.0 cm + 0.25cm,1.5cm) {\footnotesize{\color{blue!40}1\color{black}982 / 1054}};
  
  \draw[fill=blue!40] (9.25 cm,0cm) rectangle (9.25cm+0.5 cm,4.5294 cm) %die Säulen
    node at (9.25cm + 0.25 cm,4.5294 cm + 0.3cm) {\small{0.91}}; %die Prozente über den Säulen
  \node[] at (9.25 cm + 0.25cm,-0.5cm) {7}; %Säulenbeschriftung
  \node[rotate=90] at (9.25 cm + 0.25cm,1.5cm) {\footnotesize{693 / 765}};
  
  \draw[fill=blue!40] (10.5 cm,0cm) rectangle (10.5cm+0.5 cm,4.2241 cm) %die Säulen
    node at (10.5cm + 0.25 cm,4.2241 cm + 0.3cm) {\small{0.84}}; %die Prozente über den Säulen
  \node[] at (10.5 cm + 0.25cm,-0.5cm) {8}; %Säulenbeschriftung
  \node[rotate=90] at (10.5 cm + 0.25cm,1.5cm) {\footnotesize{539 / 638}};
  
  \draw[fill=blue!40] (11.75 cm,0cm) rectangle (11.75cm+0.5 cm,4.34275 cm) %die Säulen
    node at (11.75cm + 0.25 cm,4.34275 cm + 0.3cm) {\small{0.87}}; %die Prozente über den Säulen
  \node[] at (11.75 cm + 0.25cm,-0.5cm) {9}; %Säulenbeschriftung
  \node[rotate=90] at (11.75 cm + 0.25cm,1.5cm) {\footnotesize{337 / 388}};
  
  \draw[fill=blue!40] (13.0 cm,0cm) rectangle (13.0cm+0.5 cm,4.0909 cm) %die Säulen
    node at (13.0cm + 0.25 cm,4.0909 cm + 0.3cm) {\small{0.82}}; %die Prozente über den Säulen
  \node[] at (13.0 cm + 0.25cm,-0.5cm) {10}; %Säulenbeschriftung
  \node[rotate=90] at (13.0 cm + 0.25cm,1.5cm) {\footnotesize{72 / 88}};
  
  \draw[fill=blue!40] (14.25 cm,0cm) rectangle (14.25cm+0.5 cm,3.7179 cm) %die Säulen
    node at (14.25cm + 0.25 cm,3.7179 cm + 0.3cm) {\small{0.74}}; %die Prozente über den Säulen
  \node[] at (14.25 cm + 0.25cm,-0.5cm) {11}; %Säulenbeschriftung
  \node[rotate=90] at (14.25 cm + 0.25cm,1.5cm) {\footnotesize{29 / 39}};
  
  \draw[fill=blue!40] (15.5 cm,0cm) rectangle (15.5cm+0.5 cm,4.3 cm) %die Säulen
    node at (15.5cm + 0.25 cm,4.3 cm + 0.3cm) {\small{0.86}}; %die Prozente über den Säulen
  \node[] at (15.5 cm + 0.25cm,-0.5cm) {12}; %Säulenbeschriftung
  \node[rotate=90] at (15.5 cm + 0.25cm,1.5cm) {\footnotesize{43 / 50}};
  
  \draw[fill=black!40!green!50] (18.5 cm,0cm) rectangle (18.5cm+0.5 cm,4.796 cm) %die Säulen
    node at (18.5cm + 0.25 cm,4.796 cm + 0.3cm) {\small{0.96}}; %die Prozente über den Säulen 
  \node[] at (18.75 cm,-0.5cm) {No};%Säulenbeschriftung
  \node[rotate=90] at (18.75 cm,1.5cm) {\footnotesize{9414 / 9814}};
  
  \draw[fill=black!40!green!50] (19.75 cm,0cm) rectangle (19.75cm+0.5 cm,4.7585 cm) %die Säulen
    node at (19.75cm + 0.25 cm,4.7585 cm + 0.3cm) {\small{0.95}}; %die Prozente über den Säulen
  \node[] at (19.75 cm + 0.25cm,-0.5cm) {Yes}; %Säulenbeschriftung
  \node[rotate=90] at (19.75 cm + 0.25cm,1.5cm) {\footnotesize{38248 / 40186}};

  \draw[fill=black!10!red!50] (22.75 cm,0cm) rectangle (22.75cm+0.5 cm,4.796 cm) %die Säulen
    node at (22.75cm + 0.25 cm,4.796 cm + 0.3cm) {\small{0.96}}; %die Prozente über den Säulen 
  \node[] at (23.0 cm,-0.5cm) {No};%Säulenbeschriftung
  \node[rotate=90] at (23.0 cm,1.5cm) {\footnotesize{36742 / 38110}};
  
  \draw[fill=black!10!red!50] (24 cm,0cm) rectangle (24cm+0.5 cm,4.6 cm) %die Säulen
    node at (24cm + 0.25 cm,4.6 cm + 0.3cm) {\small{0.92}}; %die Prozente über den Säulen
  \node[] at (24 cm + 0.25cm,-0.5cm) {Yes}; %Säulenbeschriftung
  \node[rotate=90] at (24 cm + 0.25cm,1.5cm) {\footnotesize{10920 / 11890}};

  \node[] at (8.25 cm,-1.3cm) {(a) Age difference in years\phantomsubcaption\label{fig:age}};
  \node[rotate=90] at (-2 cm,3cm) {True Positive Rate};
  
  \node[] at (19.375 cm,-1.3cm) {(b) Abnormality\phantomsubcaption\label{fig:abnorm}};
   \node[] at (19.375 cm,-1.9cm) {change};
  
  \node[] at (23.625 cm,-1.3cm) {(c) View\phantomsubcaption\label{fig:view}};
  \node[] at (23.625 cm,-1.9cm) {change};

\end{tikzpicture}
}
        \caption{\Acp{TPR} for image pairs with \subref{fig:age} age differences, \subref{fig:abnorm} with changes in the disease pattern, and~\subref{fig:view} with changes in the projection view. The absolute numbers of true positives and overall positives are given for each bin. Note that the number of image pairs with age differences of more than 12 years is comparatively small, which is why the corresponding TPRs are neglected in this figure.}
        \label{fig:age_disease_robustness}
    \end{figure}
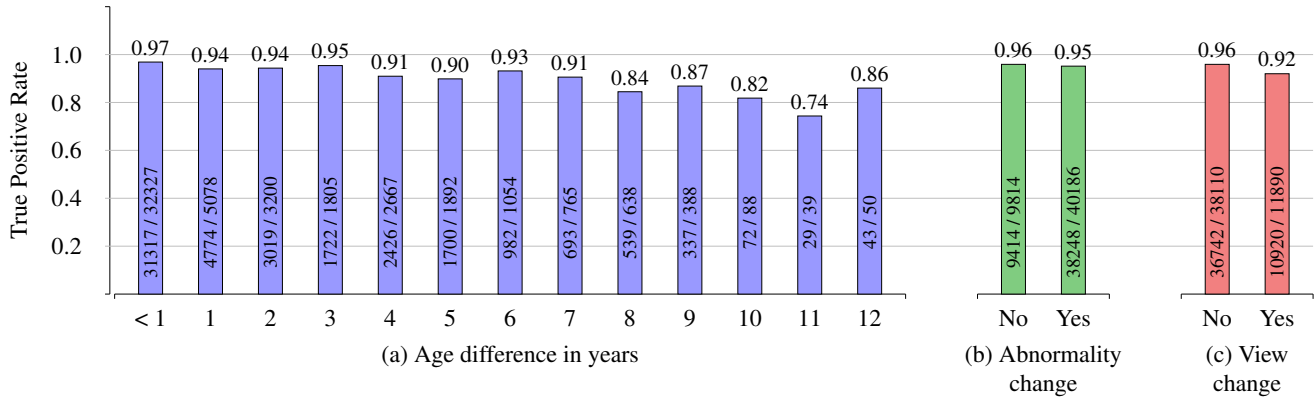
\newline
\indent
We also analyzed how the model with the best recall (fourth row in \cref{tab:results}) behaves when comparing images of the same patient where the acquisition dates are several years apart. The results are illustrated in \cref{fig:age}. We received a \ac{TPR} of 0.97 for image pairs that had small age differences of less than one year. As the age variation between the follow-up images and the initial scan increases, we observe a slight decrease in the \ac{TPR} values. Nevertheless, our model still shows competitive results even if the patient's age in two images differs by several years. 
Even for an age difference of twelve years, we can verify that two images belong to the same patient by \SI{86}{\percent}. We only report the \acp{TPR} for image pairs with follow-up intervals of up to 12 years in \cref{fig:age} as the number of pairs with larger intervals is relatively small. 
    \begin{figure}
    \centering
    \def\svgwidth{\textwidth}
    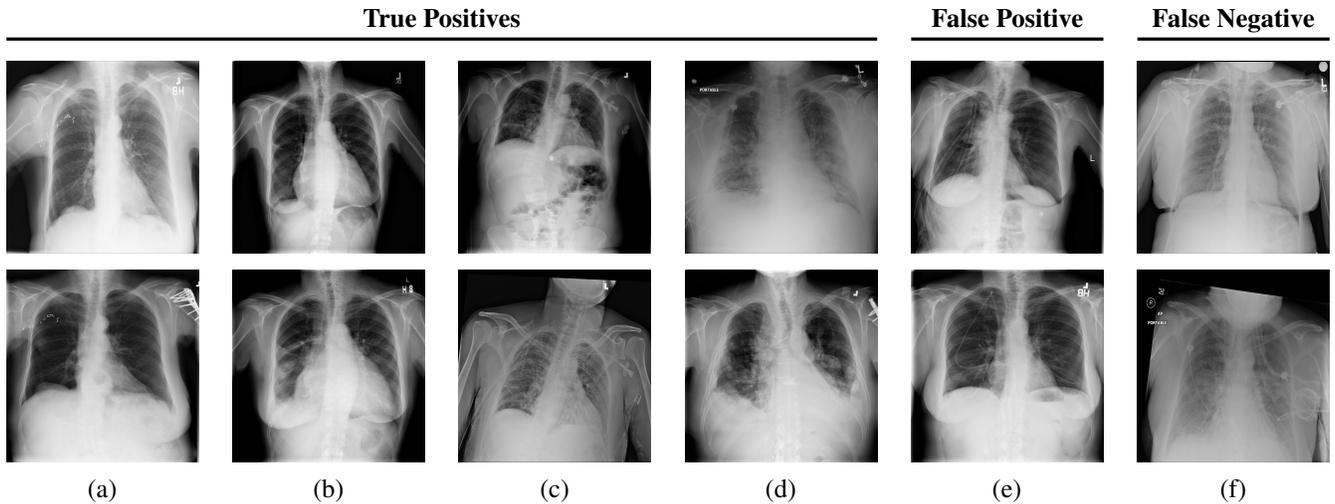
    \caption{Exemplary image pairs are classified by our best performing verification model. Each column represents one image pair. The first four columns \subref{fig:FP1}--\subref{fig:TP3} show true positive classifications. The last two columns \subref{fig:TP4} and \subref{fig:TP2} depict a false positive and a false negative classification, respectively.}
    \label{fig:image_pairs2}
    \end{figure}
    
\begin{figure}[tb]
    \centering
    \def\svgwidth{\textwidth}
    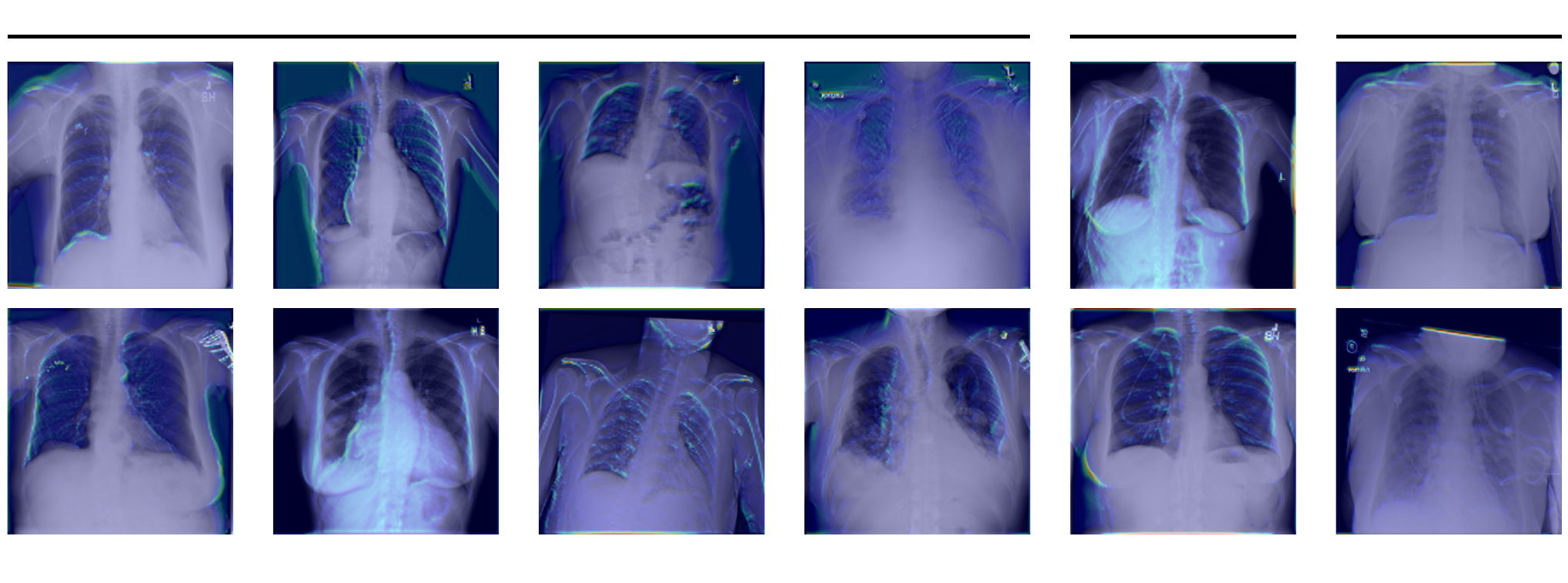
    \caption{Grad-CAM visualizations for the first convolutional layer of the ResNet-50 incorporated in our \ac{SNN}. Each column represents one image pair. The first four columns \subref{fig:FP1_map1}--\subref{fig:TP3_map1} show true positive classifications. The last two columns \subref{fig:TP4_map1} and \subref{fig:TP2_map1} depict a false positive and a false negative classification, respectively. The shown images illustrate that the anatomical structure of, e.\,g. the breast (cf.\ \subref{fig:FP1_map1}, \subref{fig:FN1_map1}, \subref{fig:TP4_map1}), the lungs (cf.\ \subref{fig:FP1_map1}, \subref{fig:FN1_map1}, \subref{fig:TP1_map1}, \subref{fig:TP4_map1}), and the heart (cf.\ \subref{fig:FP1_map1}, \subref{fig:FN1_map1}) have a high impact on the final model prediction. Furthermore, it can be seen that our network focuses on the collarbones (cf.\ \subref{fig:FP1_map1}, \subref{fig:TP1_map1}, \subref{fig:TP3_map1}, \subref{fig:TP2_map1}) and the ribs (cf.\ \subref{fig:FN1_map1}, \subref{fig:TP1_map1}, \subref{fig:TP4_map1}). The upper images of \subref{fig:FP1_map1} and \subref{fig:TP2_map1} also highlight that our network pays attention to the contour of the diaphragm.}
    \label{fig:heatmaps}
\end{figure}

\begin{figure}[t!]
    \centering
    \def\svgwidth{\textwidth}
    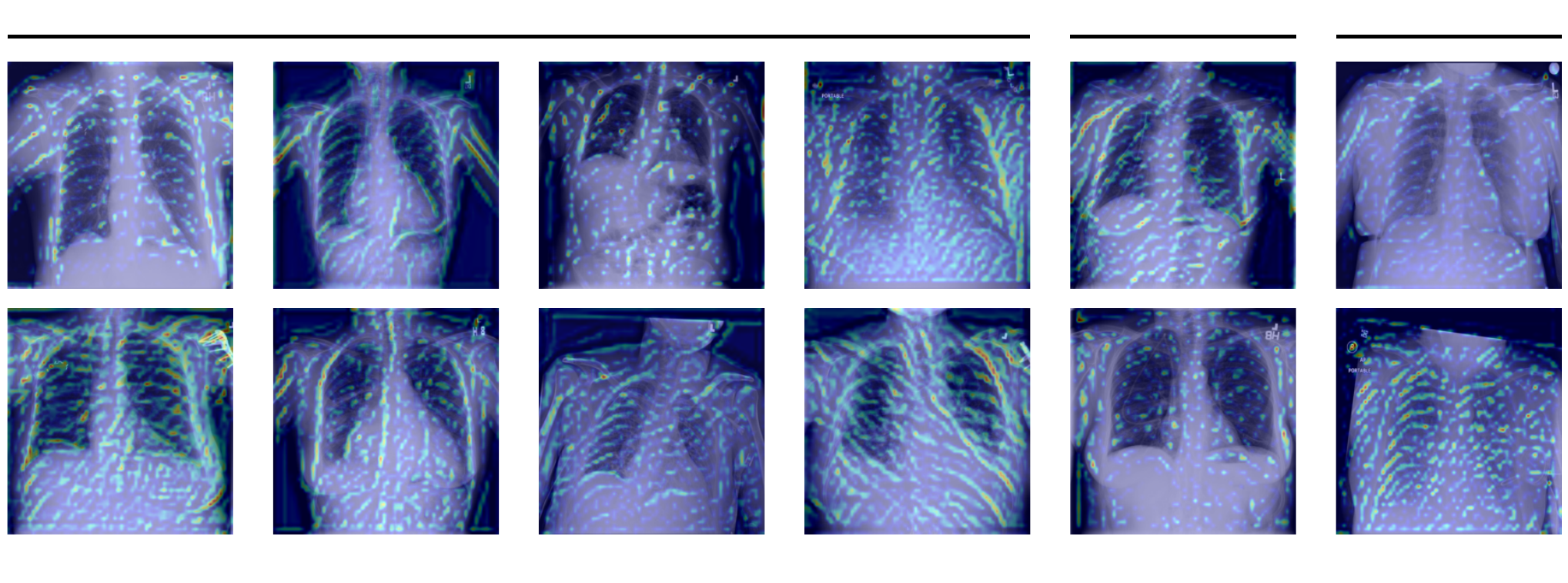
    \caption{Grad-CAM visualizations for an intermediate convolutional layer of the ResNet-50 incorporated in our \ac{SNN}. Each column represents one image pair. The first four columns \subref{fig:FP1_map3}--\subref{fig:TP3_map3} show true positive classifications. The last two columns \subref{fig:TP4_map3} and \subref{fig:TP2_map3} depict a false positive and a false negative classification, respectively. The obtained attention maps clearly illustrate that the selected network layer focuses on the ribs and the outline of the thorax.}
    \label{fig:heatmaps3}
\end{figure}

Additionally, we investigated the model's verification capability in the case of new abnormality patterns appearing in follow-up scans that did not occur in previously acquired chest radiographs. 
\Cref{fig:abnorm} shows that regardless of the abnormality, we nearly observe no decline in the \ac{TPR} values, emphasizing the robustness of our trained \ac{SNN} architecture. Note that the disease labels in the ChestX-ray14 dataset were extracted using natural language processing techniques. This could potentially have caused label noise which could have affected the results shown in \Cref{fig:abnorm}. Furthermore, \Cref{fig:view} illustrates that changes in the projection view (e.\,g., one image taken in the anterior-posterior position and the other image acquired using the posterior-anterior view) hardly lead to any deterioration in the performance.
\begin{table}[tb]
   \centering
   \caption{Comparison of the verification performance on two different subsets of the ChestX-ray14 dataset that either contain foreign material or not (first two rows). Furthermore, we show the verification results for the CheXpert dataset and the COVID-19 Image Data Collection (last two rows). We present the \ac{AUC} (together with the lower and upper bounds of the \SI{95}{\percent} confidence intervals from 10,000 bootstrap runs), the accuracy, the specificity, the recall, the precision, and the F1-score.}
   \resizebox{\textwidth}{!}{
   \renewcommand{\arraystretch}{1.3}
\begin{tabular}{|c|c|c|c|c|c|c|c|}
\hline
\rowcolor[HTML]{C0C0C0} 
\textbf{Dataset} & \textbf{Subset}         &  \textbf{AUC + 95\,\% CI} & \textbf{Accuracy ($\frac{TP+TN}{P+N}$)} & \textbf{Specificity ($\frac{TN}{N}$)} & \textbf{Recall ($\frac{TP}{P}$)} & \textbf{Precision ($\frac{TP}{TP+FP}$)} & \textbf{F1-score} \\ \hline
& w/ foreign material          & $0.9970\enspace_{0.9938}^{0.9993}$          & $0.9796\enspace (\frac{672}{686})$         & $0.9854\enspace (\frac{338}{343})$               & $0.9738\enspace (\frac{334}{343})$ & $0.9853\enspace (\frac{334}{339})$ & $0.9795$              \\ \cline{2-8}
\multirow{-2}{*}{ChestX-ray14} & w/o foreign material          & $0.9972\enspace_{0.9909}^{0.9999}$            & $0.9862\enspace (\frac{430}{436})$         & $0.9908\enspace (\frac{216}{218})$               & $0.9817\enspace (\frac{214}{218})$ & $0.9907\enspace (\frac{214}{216})$ & $0.9862$               \\ \hline
CheXpert & -          & $0.9870\enspace_{0.9855}^{0.9884}$            & $0.9440\enspace (\frac{15,562}{16,486})$         & $0.9629\enspace (\frac{7,937}{8,243})$               & $0.9250\enspace (\frac{7,625}{8,243})$ & $0.9614\enspace (\frac{7,625}{7,931})$ & $0.9429$               \\ \hline
COVID-19 & -          & $0.9763\enspace_{0.9696}^{0.9825}$            & $0.9180\enspace (\frac{1,421}{1,548})$         & $0.9780\enspace (\frac{757}{774})$               & $0.8579\enspace (\frac{664}{774})$ & $0.9750\enspace (\frac{664}{681})$ & $0.9127$               \\ \hline
\end{tabular}
   }
   \label{tab:verification_subsets_table}
\end{table}
%\newpage
\newline
\indent
Moreover, we perform a qualitative evaluation where we visually inspect some exemplary image pairs evaluated using our best-performing verification model. In \cref{fig:image_pairs2}, we show four \ac{TP} classifications \subref{fig:FP1}--\subref{fig:TP3}, one pair that has been classified as a \ac{FP} \subref{fig:TP4}, and one example for a \ac{FN} image pair \subref{fig:TP2}. The shown images clearly illustrate the high technical variance present in the ChestX-ray14 dataset. The first image pair~\subref{fig:FP1} shows two images belonging to the same patient with a difference of seven years. Clear differences in pixel intensities and lung shape are observed. 
However, both images belong to the same person, cf.\ the small vascular clips in the area of the upper right lung.
Also, image pairs with large difference in scaling \subref{fig:FN1} or rotation \subref{fig:TP1} are verified correctly. 
Our model is also robust to the patients' pathology: While the upper image of \subref{fig:FN1} shows characteristics of pneumothorax, the patient suffered from cardiomegaly, effusion, and masses in the lower image, according to the provided annotations. Similarly in \subref{fig:TP1}, where the upper image indicates the presence of infiltration and pneumothorax, whereas the lower scan shows signs of infiltration and nodules.
\Cref{fig:TP4} shows an exemplary image pair that has falsely been classified as positive. Conversely, \subref{fig:TP2} depicts a positive image pair that has been incorrectly classified as negative. To visually demonstrate which parts of the images are responsible for the verification task, we applied a siamese attention mechanism~\cite{zheng2019re} to our network architecture which utilizes the Grad-CAM algorithm~\cite{selvaraju2017grad}. The obtained attention maps can be seen in Figures~\ref{fig:heatmaps} and~\ref{fig:heatmaps3}. They clearly indicate that the human anatomy, especially the shape of the lungs and ribs, is the driving factor for the network decisions.

To investigate how foreign material (see \cref{fig:FP1}) affects the verification performance, we evaluated our trained network on two small manually created subsets of around 200 images. The first one consisted only of images in which foreign material is visible, whereas the second one solely contained images without foreign material. When constructing the subsets, we selected the patients at random and then assigned the corresponding patient images to the respective subset after visual assessment. Furthermore, we ensured that no more than 5 images were used per patient. \Cref{tab:verification_subsets_table} summarizes the results indicating that the patient verification works with high performance regardless of the occurrence of foreign material. We even observe a slight improvement in performance for the subset where no foreign material is visible in the images.

Finally, to analyze whether our trained model is able to generalize to other datasets which have not been used during training, we evaluated our network on external datasets such as CheXpert and the COVID-19 Image Data Collection. For this, we utilized 16,486 image pairs from the CheXpert dataset and 1,548 image pairs from the COVID-19 Image Data Collection. The results are summarized in the last two rows of \cref{tab:verification_subsets_table}. It can be seen that our network still yields high \ac{AUC} values of 0.9870 (CheXpert) and 0.9763 (COVID-19) for the verification task although the model has not been fine-tuned on the respective datasets. Also the other presented evaluation metrics show competitive values without deteriorating too much.  

\section*{Patient Re-identification}
For our patient re-identification experiments, we trained another \ac{SNN} architecture on the ChestX-ray14 dataset. In contrast to the verification model, we omitted all the layers from the merging layer onwards.
The main objective was to learn appropriate feature representations instead of directly determining whether the inputs belong to the same patient or not.
After training the network, we used the ResNet-50 backbone as a feature extractor for the actual image retrieval task. By computing the Euclidean distance between the embeddings of the query image and each other image, we obtained for each query image a ranked list of its most similar images in terms of identity. The used training, validation, and testing set consisted of 61,755, 10,815, and 25,596 images, respectively.
    \begin{table}[tb]
        \centering
        \caption{Overview of the obtained results for our image retrieval experiments. In this table, we report the $mAP@R$, the $R$-Precision, and the Precision@1. The first 4 rows show the results on the ChestX-ray14 dataset for different image resolutions used for evaluation. The fifth row shows the outcomes on the CheXpert dataset.
        The last row indicates the results on the COVID-19 Image Data Collection. Bold text represents the overall highest performance metrics.}
        \renewcommand{\arraystretch}{1.3}
\begin{tabular}{|c|c|c|c|c|}
\hline
\rowcolor[HTML]{C0C0C0} 
\textbf{Dataset}               & \textbf{Input dimensions} & \textbf{mAP@R} & \textbf{R-Precision} & \textbf{Precision@1} \\ \hline
                 & 1024$\times$1024 & \textbf{0.9748} & \textbf{0.9763} & \textbf{0.9963} \\ \cline{2-5} 
                 & 800$\times$800   & 0.9709 & 0.9726 & 0.9958 \\ \cline{2-5} 
                 & 512$\times$512   & 0.9572 & 0.9601 & 0.9945 \\ \cline{2-5} 
\multirow{-4}{*}{ChestX-ray14} & 224$\times$224            & 0.7730         & 0.7979               & 0.9756               \\ \hline
CheXpert & original         & 0.8001 & 0.8148 & 0.9444 \\ \hline
COVID-19 & original         & 0.8569 & 0.8707 & 0.8821 \\ \hline
\end{tabular}

        \label{tab:results_retrieval}
    \end{table}
    \begin{table}[tb]
        \centering
        \caption{Comparison of the re-identification performance on two different subsets (ChestX-ray14) that either contain foreign material or not. We report the $mAP@R$, the $R$-Precision, and the Precision@1.}
        \renewcommand{\arraystretch}{1.3}
\begin{tabular}{|c|c|c|c|}
\hline
\rowcolor[HTML]{C0C0C0} 
\textbf{Subset}                & \textbf{mAP@R} & \textbf{R-Precision} & \textbf{Precision@1} \\ \hline
                w/ foreign material  & 0.9925 & 0.9925 & > 0.9999 \\ \hline
                w/o foreign material & > 0.9999 & > 0.9999 & > 0.9999 \\ \hline
\end{tabular}
        \label{tab:results_table_retrieval_subsets}
    \end{table}
\newline
\indent
The results of the corresponding image retrieval experiments are summarized in \cref{tab:results_retrieval}. When using the original image size of 1024$\times$1024 pixels for evaluation, we obtain a precision@1 of more than \SI{99}{\percent} showing that the closest match nearly always is the same patient. The high mean average precision at $R$ (mAP@$R$) of about \SI{97}{\percent} further depicts that most of the most similar images are correctly identified. We observe a slight decrease in performance as the image size was reduced. Nevertheless, when the images were downsampled to a resolution of 512$\times$512 pixels, we still obtain high performance values. When the image size was reduced too aggressively, e.\,g., to 224$\times$224 pixels, the $mAP@R$ and the $R$-Precision rates drop. Yet, we still observed a high Precision@1 of more than \SI{97}{\percent}.
\newline
\indent
Similar to the experiments in the patient verification section, we evaluated our best-trained re-identification model on two small subsets, one of which only contained images with visible foreign material and the other consisted exclusively of images without the presence of foreign material. The obtained results are presented in \cref{tab:results_table_retrieval_subsets}. As can be seen, we achieve high performance values for both subsets. Thus, we hypothesize that our outcomes are independent of foreign material which may occur only for specific patients. 
\newline
\indent
Lastly, we analyzed the re-identification performance on external datasets such as CheXpert and the COVID-19 Image Data Collection. For this, we utilized 6,454 images from the CheXpert dataset and 781 images from the COVID-19 dataset. As can be seen in the last two rows of \cref{tab:results_retrieval}, we also  obtain high retrieval values although we haven't performed any fine-tuning on both datasets, which demonstrates the feasibility of the trained re-identification network on previously unseen datasets.

%
% DISCUSSION
%
\section*{Discussion and Conclusion}
In this paper, we investigated the patient verification and re-identification capabilities of \ac{DL} techniques on chest radiographs. We have shown that well-trained \ac{SNN} architectures are able to compare two individual frontal chest radiographs and reliably predict whether these images belong to the same patient or not. Moreover, we have shown that \ac{DL} models have the potential to accurately retrieve relevant images in a ranked list. Our models have been evaluated on the publicly available ChestX-ray14 dataset and showed competitive results with an \ac{AUC} of up to 0.9940 and classification accuracy of more than 95\,\% in the verification scenario and an $mAP@R$ of \SI{97}{\percent} and a precision@1 of about \SI{99}{\percent} in the image retrieval scenario. Especially the fact that basic \acp{SNN} have the capability to re-identify patients despite potential age differences, disease changes or differing projection views demonstrated the effectiveness of \ac{DL} techniques for this task. However, note that the shown results were obtained empirically, i.\,e.\ they do not necessarily reflect true measures of certainty.

As shown in \cref{fig:image_pairs2}, the used dataset suffers from a high technical variance which may occur due to various windowing techniques applied to the images. In a real-life scenario, the resulting variations in image contrast and brightness could be significantly mitigated by using dynamic normalization approaches~\cite{gundel2021robust}. Furthermore, we believe that variations in rotation and scaling can be counteracted by appropriate alignment algorithms. Nevertheless, even without such pre-processing steps, we were able to show that patient matching for chest radiographs is possible with a high performance by using \ac{DL} techniques.

Moreover, we hypothesize that special noise patterns characteristic for unique patients appear in the images which might unintentionally improve the re-identification performance. For example, the initial anonymization strategy may be biased towards the clinical institution and, therefore, also towards follow-up images. To get a better impression of the re-identification capability of our \ac{SNN} architecture, we also intend to investigate other datasets which show less or ideally no correlation between potential noise patterns and the patient identity. Therefore, further research on multiple datasets should ideally be considered. For our experiments, we already evaluated our models on two completely different datasets, the CheXpert dataset and the COVID-19 Image Data Collection. While the evaluation metrics are lower (possibly due to a domain shift or the severity of diseases), we still obtain \ac{AUC} scores of over \SI{97}{\percent} (COVID-19) and \SI{98}{\percent} (CheXpert) and precision@1 values of more than \SI{88}{\percent} (COVID-19) and \SI{94}{\percent} (CheXpert) without fine-tuning on these datasets. This indicates that patient verification and re-identification is also applicable for data that was acquired in various hospitals around the world where other pre-processing steps may be taken before data publication compared to the ChestX-ray14 dataset.

The COVID-19 Image Data Collection is very heterogeneous, containing, e.\,g., images of different sizes, both gray-scale and color images, and images with visible markers, arrows or date displays. For our experiments, only those images in the COVID-19 Image Data Collection were used that were acquired using the anterior-posterior or the posterior-anterior view, while images taken in the lateral position and CT scans were discarded. Apart from this, no further steps were taken to ensure the quality of the dataset. Although some of the factors mentioned above (e.\,g., brandings such as markers, arrows or dates) may facilitate the patient re-identification, we hypothesize that the COVID-19 Image Data Collection poses a realistic example of a public medical dataset and we therefore consider the conducted experiment as an authentic real-life application scenario.

Furthermore, we want to accentuate that our trained network architectures are able to handle non-rigid transformations that may appear between two images of the same person in the ChestX-ray14 dataset. Such deformations can occur due to various breath states in follow-up scans or due to different positioning during X-ray acquisition. Hence, the shape of the heart and lungs, or the contours of the ribs may appear deformed compared to an initial scan. The obtained results lead to the assumption that our trained \ac{SNN} architectures can withstand such deformations and can therefore be used for reliable patient re-identification on chest radiographs.

We conclude that \ac{DL} techniques render medical chest radiographs biometric for everyone and allow a re-identification similar to a fingerprint. Therefore, publicly available medical chest X-ray data is not entirely anonymous. Using a \ac{DL}-based re-identification network enables an attacker to compare a given radiograph with public datasets and to associate accessible metadata with the image of interest. The strength of our proposed method is that patients can be re-identified in a fully automated way without the need of expert knowledge. Thus, sensitive patient data is exposed to a high risk of falling into the unauthorized hands of an attacker who may disseminate the gained information against the will of the concerned patient. At this point, we want to emphasize that data leakage of this kind requires that the attacker has previously gained access to an image of a known person. This could happen, for example, through a stolen CD containing raw medical data of a specific patient, or through accidental data release by a radiological facility. Furthermore, data breaches due to inadequate data security measures at, e.\,g., healthcare institutions or health insurance companies, represent a possibility for attackers to obtain images of known patients, which could subsequently be utilized for a linkage attack as presented in our work. However, even if the attacker owns an image of an unknown identity, a re-identification model can be used to find the same patient across various datasets. Assuming multiple datasets contain the same patient but different metadata, an attacker would be able to obtain a more complete picture of the respective patient. We hypothesize that collecting patient information by this means could significantly help an attacker infer the true identity of the patient. We therefore urge that conventional anonymization techniques be reconsidered and that more secure methods be developed to resist the potential attacks by DL-based algorithms.

At this point, we would like to draw attention to the analogy of our work to the field of \ac{ASV}. Speech signals contain a large amount of private data, e.\,g., age, gender, health and emotional state, ethnic origin, and more~\cite{nautsch2019preserving}. As such information is embedded in speech data itself, it can be exploited to reveal the speaker's identity by applying attack models in the form of \ac{ASV} systems~\cite{nautsch2019preserving}. Therefore, in the speech community, raw speech signals are not considered anonymous. Instead, privacy challenges, such as the VoicePrivacy Challenge~\cite{vpc2022}, were formed to develop solutions for the preservation of privacy in this field. With our work, we were able to demonstrate that the privacy issue with speech data is 1-to-1 transferable to the privacy issue with chest radiographs.

While the proposed algorithms may be used maliciously to produce harm in terms of patient privacy and data anonymity, we also want to draw attention to a potentially positive application area. Often, different datasets are used for training and evaluation of an algorithm. However, since in most cases these datasets have been anonymized using conventional techniques, it is not clear whether certain patients appear in more than one dataset. Therefore, in this context, our trained networks could be applied to check for mutual exclusiveness with respect to included patients between multiple datasets.

The publication of medical image datasets is an area of conflict. While, on the one hand, many patients may benefit from recent advances (e.\,g., the development of diagnostic algorithms), there are, on the other hand, patients who may be seriously harmed by the fact that their data is publicly available. With our work, we focus on providing empirical evidence for this issue and draw attention to the risks. The legal situation for the publication of medical data is currently regulated by the HIPAA (in the United States) and the GDPR (in Europe). We therefore contend that the corresponding ethics committees are responsible for weighing the benefits and the risks as well as for assessing the appropriateness of current regulations.

Potential solutions to the problems addressed in our work may be found in privacy-preserving approaches such as \ac{DP}~\cite{dwork2011firm,dwork2014algorithmic} which guarantees that the global statistical distribution of a dataset is retained while individually recognizable information is reduced~\cite{kaissis2020secure}. This means that an outside observer is unable to draw any conclusions about the presence or absence of a particular individual. Consequently, algorithms trained with \ac{DP} are able to withstand linkage attacks attempting to reveal the identities of patients in the dataset used to train the algorithm. One commonly-used technique to achieve \ac{DP} is to modify the input by adding noise to the dataset (local \ac{DP})~\cite{kaissis2020secure,sarwate2013signal}. Furthermore, \ac{DP} can be applied to the computation results (global \ac{DP}) or to algorithm updates~\cite{kaissis2020secure}. However, training models with \ac{DP} degrades the quality of the model (privacy-utility trade-off) which is problematic in medicine where high diagnostic utility is required. Therefore, further exploration on these topics is necessary before general conclusions can be made.

Aside from perturbation-based privacy approaches, we want to mention that the use of collaborative decentralized learning protocols such as \ac{FL}~\cite{konevcny2016federated} can significantly contribute to a safer use of medical data. By training a machine learning model collaboratively without centralizing the data, the need of raw data sharing or dataset release is eliminated~\cite{rieke2020future}. Thus, the medical data is able to reside with its owner, e.\,g., the healthcare institution where the data was acquired, which resolves data governance and ownership issues~\cite{kaissis2020secure,kaissis2021end}. However, \ac{FL} itself does not provide full data security and privacy, meaning that some risks remain unless combined with other privacy-preserving methods.

%
% METHODS
%
\section*{Methods}
The research was carried out in accordance with the relevant guidelines and regulations of the institution conducting the experiments.
\subsection*{Siamese neural networks}
To re-identify patients from their chest radiographs, we employ \ac{SNN} architectures for both the classification and the retrieval tasks. A \ac{SNN} receives two input images which are processed by two identical feature extraction blocks sharing the same set of network parameters. The resulting feature representations can then be used to compare  the inputs. The concept of a \ac{SNN} was initially introduced by Bromley et al.~\cite{bromley1993signature} for handwritten signature verification. Taigman et al.~\cite{taigman2014deepface} applied this idea in the field of face verification and proposed the \textit{DeepFace} system. Moreover, Koch et al.~\cite{koch2015siamese} presented an approach for one-shot learning on the Omniglot~\cite{lake2015human} and MNIST~\cite{lecun1998gradient} datasets.

\subsection*{NIH ChestX-ray14 dataset}
With a total of 112,120 frontal-view chest radiographs from 30,805 unique patients, the NIH ChestX-ray14~\cite{wang2017chestx} dataset counts to one of the largest publicly available chest radiography datasets in the scientific community. Due to follow-up scans, the image collection provides an average of 3-4 images per patient. The originally acquired radiographs were published as 8-bit gray-scale PNG images with a size of 1024$\times$1024 pixels. Associated metadata is available for all images in the dataset. The additional data comprises information about the underlying disease patterns (either no finding or a combination of up to 14 common thoracic pathologies), the number of follow-up images taken, the patients' age and gender, and the projection view (anterior-posterior or posterior-anterior) used for radiography acquisition. According to the publisher, the dataset was carefully screened to remove all personally identifiable information before release~\cite{NIHClinicalCenter}. Therefore, the patient names were replaced by integer IDs. Moreover, personal data in the image domain itself has been made unrecognizable by placing black boxes over the corresponding image areas.

\subsection*{CheXpert dataset}
The CheXpert~\cite{irvin2019chexpert} dataset contains 224,316 frontal and lateral chest radiographs of 65,240 patients, who underwent a radiographic examination from Standford University Medical Center between October 2002 and July 2017. The originally acquired radiographs were published as 8-bit gray-scale JPG images with varying image resolutions. Note that only frontal chest radiographs were used in our work, whereas lateral images were excluded. 

\subsection*{COVID-19 Image Data Collection}
The COVID-19 Image Data Collection~\cite{cohen2020covid} is a dataset that was created and published as an initiative to provide COVID-19 related chest radiographs and CT scans for machine learning tasks. It comprises data of 448 unique patients and a total of around 950 images with different image resolutions. In this work, only the available frontal radiographs were utilized, whereas the lateral images and CT scans were discarded. 

\subsection*{Data preparation}
Since \ac{SNN} architectures require pairs of images for training and evaluation, we construct both positive and negative image pairs from the images contained in the ChestX-ray14 dataset. In this context, a positive pair consists of two images belonging to the same patient, whereas a negative pair comprises two images that belong to different patients. Mathematically, the constructed dataset can be described according to
\begin{linenomath*}
\begin{equation}
    \mathcal{S} = \{(\bm{x}_{11}, \bm{x}_{12}, y_1), ..., (\bm{x}_{m1}, \bm{x}_{m2}, y_m), ..., (\bm{x}_{M1}, \bm{x}_{M2}, y_M) \}\enspace,\enspace \text{with } y_m \in \{0,1\}\enspace,
\end{equation}
\end{linenomath*}
where the triplet $(\bm{x}_{m1}, \bm{x}_{m2}, y_m)$ represents one sample consisting of two images $\bm{x}_{m1}$ and $\bm{x}_{m2}$, and the corresponding label $y_m$. $M$ refers to the total number of samples and $m$ denotes an iterator variable in the range of $m \in [1,M]$. The class label $y_m$ symbolizes a binary variable that takes the value 0 for negative image pairs and 1 for positive image pairs.

To ensure that images from one patient only appear either in the training, validation, or testing set, we use the patient-wise splitting strategy. According to the official split provided with the ChestX-ray14 dataset, the data is randomly divided into 70\,\% training, 10\,\% validation, and 20\,\% testing. Based on this split, we construct the actual image pairs for each subset. 

\subsubsection*{Offline mining}
For patient verification, we follow an offline mining approach, meaning that the positive and negative image pairs are generated once before conducting the experiments. First, the positive pairs are generated by only considering the patients for whom multiple images exist in the respective subset. For each patient with follow-up images, we produce all possible tuple combinations assuming the images to be unique. By following this approach, we are able to construct a total of around 400,000 positive image pairs for our training set. The negative pairs in each subset are randomly generated and concatenated with the respective positive pairs afterwards.

\subsubsection*{Online mining}
For the patient re-identification experiments, we choose an online mining approach, meaning that image pairs are formed in each batch during the training procedure. This means that the embeddings of all batch images are first computed and then subsequently used in all possible combinations as input for the loss function. Moreover, all patients with only one available image were discarded from the training set.

\subsection*{Patient verification}
\subsubsection*{Deep learning architecture}
\begin{figure}
    \centering
    \includegraphics[width=0.8\textwidth]{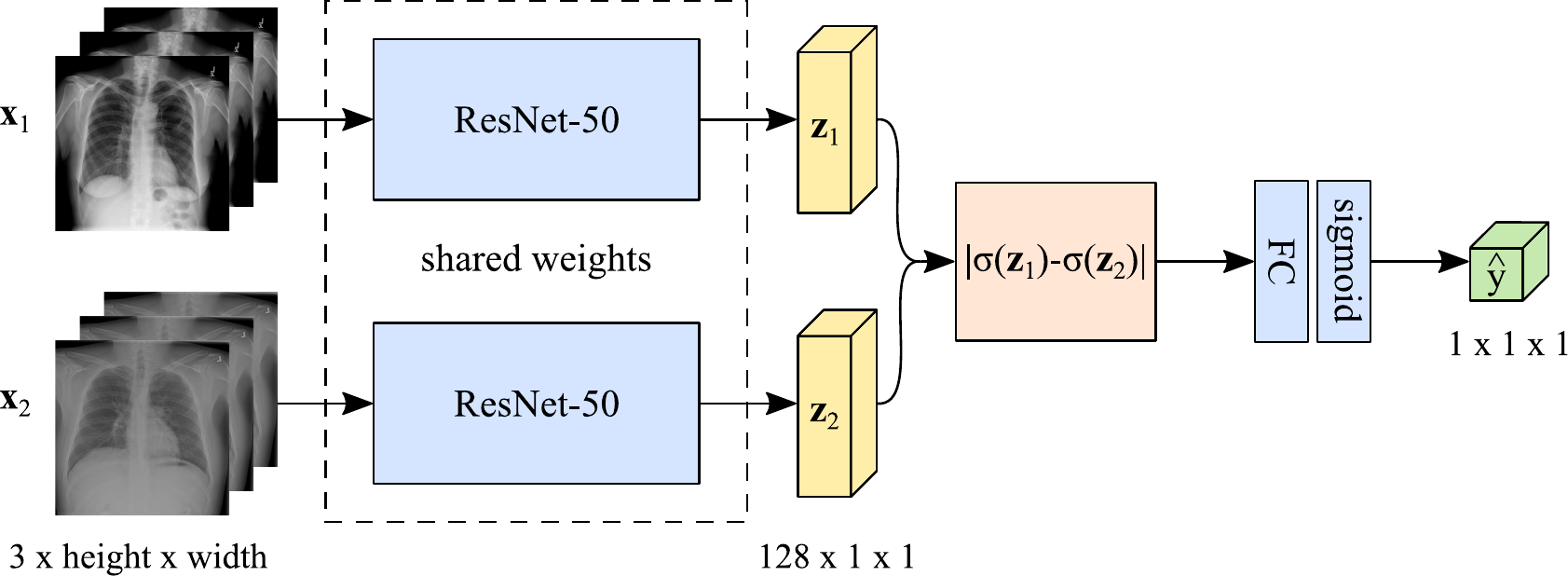}
    \caption{SNN architecture used for patient verification on the ChestX-ray14~\cite{wang2017chestx} dataset. The feature extraction blocks (light blue) share the same set of network parameters and produce the feature representations $\bm{z}_1$ and $\bm{z}_2$ (yellow). After merging (orange) and an additional FC and sigmoid layer $\sigma$ (light blue), the network yields the final output score $\hat{y}$ (green). For our patient re-identification experiments, we used the same architecture but ejected all the layers from the merging layer onwards.
    }
    \label{fig:network_architecture}
\end{figure}

For patient verification, the used \ac{SNN} architecture (see \cref{fig:network_architecture}) receives two images $\bm{x}_1$ and $\bm{x}_2$ of size 3$\times$256$\times$256. Both inputs are processed by a pre-trained ResNet-50 incorporated in each network branch. In its original version, the ResNet-50 was designed to classify images into 1,000 object categories trained on the ImageNet~\cite{deng2009imagenet} dataset. To adapt the ResNet-50 to our specific needs, we replace its classification layer with a layer consisting of 128 output neurons producing the feature representations $\bm{z}_1$ and $\bm{z}_2$, respectively. To merge both network branches, the absolute difference of the sigmoid activations of the two feature vectors is computed. We add a \ac{FC} layer to reduce the dimensionality to one neuron, followed by another sigmoid activation function $\sigma$ which yields the final output score~$\hat{y} \in [0,1]$.

\subsubsection*{Training strategy}
The verification model is trained using the \ac{BCE} loss. The network parameters are optimized by combining mini-batch \ac{SGD}~\cite{lecun2012efficient,goodfellow2016deep} with the \ac{Adam}~\cite{kingma2014adam} method. The batch size $N_b$ is set to 32 in all our experiments. We use different \acp{LR} to investigate their effect on the model's performance. Furthermore, we include an early stopping criterion with a patience $p=5$, which means that the training procedure stops as soon as the network does not improve for 5 epochs. We train the architecture using input dimensions of 3$\times$256$\times$256.

\subsubsection*{Evaluation techniques}
We utilize \ac{ROC} curves to visualize the trained verification models based on their performance. A \ac{ROC} curve represents a two-dimensional graph in which the \ac{TPR} is plotted against the \ac{FPR} at various threshold settings~\cite{fawcett2006introduction}, thus indicating how many true positive classifications can be gained as an increasing number of false positive classifications is allowed. Additionally, we calculate the \ac{AUC} which reflects a proportion of the area of the unit square and will always range from 0 to 1~\cite{fawcett2006introduction}. The higher the \ac{AUC} score, the better the model's average performance. Nevertheless, it has to be mentioned that a classifier with a high \ac{AUC} might perform worse in a specific region of \ac{ROC} space than a classifier with a low \ac{AUC} value. Moreover, we evaluate the performance by computing the accuracy, specificity, recall, precision and F1-score. Therefore, the threshold at the output neuron is set to $t=0.5$.

\subsection*{Patient re-identification}
\subsubsection*{Deep learning architecture}
For patient re-identification, we train a \ac{SNN} architecture which receives two images $\bm{x}_1$ and $\bm{x}_2$ of size 3$\times$1024$\times$1024. Both inputs are processed by a pre-trained ResNet-50 incorporated in each network branch. However, the network head of the used ResNet-50 is slightly modified. The average pooling layer is replaced by an adaptive average pooling layer producing feature maps of size 5$\times$5. In addition to the adaptive average pooling layer, an adaptive max-pooling layer is applied which also yields feature maps of size 5$\times$5. The outputs of the pooling layers are concatenated and processed by a 1$\times$1 convolutional layer reducing the number of feature maps from 2048 to 100. The feature maps are then flattened, followed by two successive \ac{FC} layers resulting in 128-dimensional feature representations $\bm{z}_1$ and $\bm{z}_2$ for the first and the second network branch. 

\subsubsection*{Training strategy}
The re-identification model is trained using the contrastive loss function~\cite{hadsell2006dimensionality} which is typically utilized to achieve a meaningful mapping $F$ from high to low dimensional space. By using the contrastive loss, the network learns to map similar inputs to nearby points on the output manifold while dissimilar inputs are mapped to distant points. Negative pairs contribute to the loss only if their distance is smaller than a certain margin $m$. In this work, the margin is set to $m=1$.
\newline
\indent
For our image retrieval experiments, the \ac{SNN} architecture is optimized using the SGD algorithm in combination with the 1cycle learning policy~\cite{smith2019super,smith2017cyclical}. When using the 1cycle \ac{LR} schedule, the \ac{LR} $\eta$ steadily increases until it reaches a chosen maximum value and gradually decreases again thereafter. This schedule changes the \ac{LR} after every single batch and is pursued a pre-defined number of epochs. 
The upper bound is chosen at 0.1584 with the help of a \ac{LR} finder. The lower bound is set to 0.0063. The L2 regularization technique is used with a decay factor of $10^{-5}$. Moreover, the batch size is adjusted to 32. We optimize the \ac{SNN} architecture by first training the adapted network head of the incorporated ResNet-50 for 30 epochs with all other parameters being frozen. Then, the complete architecture is trained for another cycle, this time consisting of 50 epochs.
\newline
\indent
Since the batch size limits the task of constructing informative positive and negative pairs in the online mining strategy, the concept of cross-batch memory~\cite{wang2020cross} is utilized to generate sufficient pairs across multiple mini-batches. This concept is based upon the observation that the embedding features generally tend to change slowly over time. This ``slow drift'' phenomenon allows the use of embeddings of previous iterations that would normally be considered out-dated and discarded. For our experiments, a memory size of 128 is chosen, meaning that the last 4 batches are considered for mining.

\subsubsection*{Evaluation techniques}
To evaluate the re-identification performance of our trained model, several metrics are computed. $R$-Precision represents the precision at $R$, where $R$ denotes the number of relevant images for a given query image. In other words, if the top-$R$ retrieved images show $r$ relevant images, then $R$-Precision can be calculated from \cref{eq:r_prec}. Note that this value is then averaged over all query samples. Precision@1 constitutes a special case and evaluates how many times the top-1 images in the retrieved lists are relevant.
\begin{linenomath*}
    \begin{equation}
        R\text{-}Precision = \frac{r}{R}\label{eq:r_prec}
    \end{equation}
\end{linenomath*}

To further consider the order of the relevant images within the retrieved list, the mean average precision at $R$ ($mAP@R$) is computed according to \cref{eq:mAP}. The $mAP@R$ denotes the mean of the average precision scores at $R$ ($AP@R$) over all $Q$ query images. The $AP@R$ (see \cref{eq:AP}) is the average of the precision values over all $R$ relevant samples, where $P@i$ refers to the precision at rank $i$ and $rel@i$ is an indicator function which equals 1 if the sample is relevant at rank $i$ and 0 if it is not relevant.

    \noindent\begin{minipage}{.5\linewidth}
        \centering
        \begin{equation}
            mAP@R = \frac{1}{Q} \sum_{i=1}^{Q} AP_{i}@R\label{eq:mAP}
        \end{equation}
    \end{minipage}%
    \begin{minipage}{.5\linewidth}
        \centering
        \begin{equation}
            AP@R = \frac{1}{R} \sum_{i=1}^{R} P@i \times rel@i \label{eq:AP}
        \end{equation}
    \end{minipage}

\subsection*{Data availability}
The NIH ChestX-ray14 dataset used throughout the current study is available via Box at \url{https://nihcc.app.box.com/v/ChestXray-NIHCC}. The COVID-19 Image Data Collection is available on GitHub at \url{https://github.com/ieee8023/covid-chestxray-dataset}. The CheXpert dataset can be requested at \url{https://stanfordmlgroup.github.io/competitions/chexpert}.

\subsection*{Code availability}
The code used to train and evaluate both the patient verification and the patient re-identification models is available at \url{https://github.com/kaipackhaeuser/CXR-Patient-ReID}. Correspondence and requests for materials should be addressed to K.P.

\subsection*{Acknowledgements}
The research leading to these results has received funding from the European Research Council (ERC) under the European Union’s Horizon 2020 research and innovation program (ERC grant no.\ 810316).

\subsection*{Author contributions}
A.M., V.C., S.G., C.S., and K.P.\ conceived the main idea. K.P.\ and N.M.\ performed the experiments and the evaluation. K.P.\ wrote the main part of the manuscript. S.G.\ offered continuous support during the experiments and the writing process. S.G., C.S., V.C., and A.M.\ provided expertise through intense discussions. All authors reviewed the manuscript.

\subsection*{Competing interests}
The authors declare no competing interests.

%
% BIBLIOGRAPHY
%
\bibliography{main}

\newpage
\section*{Supplementary Material}
\captionsetup[table]{name=Supplementary Table}
\setcounter{table}{0}
\begin{table}[H]
    \centering
    \caption{Additional patient verification results: Comparison of different data handling techniques (FTS and RNP), training set sizes $N_s$ and learning rates $\eta$. We present the AUC (together with the lower and upper bounds of the 95\,\% confidence intervals from 10,000 bootstrap runs), the accuracy, the specificity, the recall, the precision, and the F1-score.}
    \label{tab:additional_results}
    \resizebox{\textwidth}{!}{
    \renewcommand{\arraystretch}{1.3}
\begin{tabular}{|c|c|c|c|c|c|c|c|c|}
\hline
\rowcolor[HTML]{C0C0C0} 
\textbf{\begin{tabular}[c]{@{}c@{}}Data\\ handling\end{tabular}}         & \textbf{N\textsubscript{s}}                & \textbf{$\eta$} & \textbf{AUC + 95\,\% CI} & \textbf{Accuracy ($\frac{TP+TN}{P+N}$)} & \textbf{Specificity ($\frac{TN}{N}$)} & \textbf{Recall ($\frac{TP}{P}$)} & \textbf{Precision ($\frac{TP}{TP+FP}$)} & \textbf{F1-score} \\ \hline
\multirow{17}{*}{FTS} & \multirow{4}{*}{100,000} & $10^{-4}$ & $0.8509\enspace_{0.8485}^{0.8532}$             & $0.7461\enspace (\frac{74,605}{100,000})$   & $0.8414\enspace (\frac{42,071}{50,000})$       & $0.6507\enspace (\frac{32,534}{50,000})$     & $0.8040\enspace (\frac{32,534}{40,463})$             & $0.7193$    \\ \cline{3-9} 
                      &                          & $10^{-5}$ & $0.7429\enspace_{0.7399}^{0.7459}$             & $0.6322\enspace (\frac{63,223}{100,000})$   & $0.7924\enspace (\frac{39,619}{50,000})$       & $0.4721\enspace (\frac{23,604}{50,000})$     & $0.6945\enspace (\frac{23,604}{33,985})$             & $0.5621$    \\ \cline{3-9} 
                      &                          & $10^{-6}$ & $0.7257\enspace_{0.7225}^{0.7289}$             & $0.6702\enspace (\frac{67,020}{100,000})$   & $0.6719\enspace (\frac{33,594}{50,000})$       & $0.6685\enspace (\frac{33,426}{50,000})$     & $0.6708\enspace (\frac{33,426}{49,832})$             & $0.6696$    \\ \cline{3-9} 
                      &                          & $10^{-7}$ & $0.7440\enspace_{0.7409}^{0.7470}$             & $0.6741\enspace (\frac{67,410}{100,000})$   & $0.7002\enspace (\frac{35,008}{50,000})$       & $0.6480\enspace (\frac{32,402}{50,000})$     & $0.6837\enspace (\frac{32,402}{47,394})$             & $0.6654$    \\ \cline{2-9} 
                      & \multirow{4}{*}{200,000} & $10^{-4}$ & $0.8886\enspace_{0.8866}^{0.8906}$             & $0.7920\enspace (\frac{79,203}{100,000})$   & $0.8547\enspace (\frac{42,736}{50,000})$       & $0.7293\enspace (\frac{36,467}{50,000})$     & $0.8339\enspace (\frac{36,467}{43,731})$             & $0.7781$    \\ \cline{3-9} 
                      &                          & $10^{-5}$ & $0.8158\enspace_{0.8132}^{0.8185}$             & $0.6758\enspace (\frac{67,582}{100,000})$   & $0.8574\enspace (\frac{42,869}{50,000})$       & $0.4943\enspace (\frac{24,713}{50,000})$     & $0.7761\enspace (\frac{24,713}{31,844})$             & $0.6039$    \\ \cline{3-9} 
                      &                          & $10^{-6}$ & $0.7615\enspace_{0.7586}^{0.7645}$             & $0.6755\enspace (\frac{67,552}{100,000})$   & $0.7448\enspace (\frac{37,242}{50,000})$       & $0.6062\enspace (\frac{30,310}{50,000})$     & $0.7038\enspace (\frac{30,310}{43,068})$             & $0.6514$    \\ \cline{3-9} 
                      &                          & $10^{-7}$ & $0.7526\enspace_{0.7495}^{0.7556}$             & $0.6819\enspace (\frac{68,194}{100,000})$   & $0.6791\enspace (\frac{33,957}{50,000})$       & $0.6847\enspace (\frac{34,237}{50,000})$     & $0.6809\enspace (\frac{34,237}{50,280})$             & $0.6828$    \\ \cline{2-9} 
                      & \multirow{4}{*}{400,000} & $10^{-3}$ & $0.9541\enspace_{0.9529}^{0.9552}$             & $0.8852\enspace (\frac{88,519}{100,000})$   & $0.8655\enspace (\frac{43,275}{50,000})$       & $0.9049\enspace (\frac{45,244}{50,000})$     & $0.8706\enspace (\frac{45,244}{51,969})$             & $0.8874$    \\ \cline{3-9} 
                      &                          & $10^{-5}$ & $0.8518\enspace_{0.8494}^{0.8542}$             & $0.7450\enspace (\frac{74,498}{100,000})$   & $0.8410\enspace (\frac{42,050}{50,000})$       & $0.6490\enspace (\frac{32,448}{50,000})$     & $0.8032\enspace (\frac{32,448}{40,398})$             & $0.7179$    \\ \cline{3-9} 
                      &                          & $10^{-6}$ & $0.7417\enspace_{0.7386}^{0.7449}$             & $0.6572\enspace (\frac{65,715}{100,000})$   & $0.7152\enspace (\frac{35,761}{50,000})$       & $0.5991\enspace (\frac{29,954}{50,000})$     & $0.6778\enspace (\frac{29,954}{44,193})$             & $0.6360$    \\ \cline{3-9} 
                      &                          & $10^{-7}$ & $0.7306\enspace_{0.7275}^{0.7337}$             & $0.6567\enspace (\frac{65,670}{100,000})$   & $0.6801\enspace (\frac{34,006}{50,000})$       & $0.6333\enspace (\frac{31,664}{50,000})$     & $0.6644\enspace (\frac{31,664}{47,658})$             & $0.6485$    \\ \cline{2-9}
                   & \multirow{6}{*}{800,000} & $10^{-3}$ & $0.9826\enspace_{0.9820}^{0.9833}$             & $0.9324\enspace (\frac{93,238}{100,000})$   & $0.9393\enspace (\frac{46,966}{50,000})$       & $0.9254\enspace (\frac{46,272}{50,000})$     & $0.9385\enspace (\frac{46,272}{49,306})$             & $0.9319$    \\ \cline{1-1} \cline{3-9} 
\multirow{2}{*}{RNP}                   &                          & $10^{-4}$ & $\textbf{0.9940}\enspace_{0.9937}^{0.9944}$ & $0.9555\enspace (\frac{95,545}{100,000})$   & $0.9822\enspace (\frac{49,111}{50,000})$       & $0.9287\enspace (\frac{46,434}{50,000})$     & $0.9812\enspace (\frac{46,434}{47,323})$             & $0.9542$ \\ \cline{3-9}
                                       &                          & $10^{-5}$   & $0.9278\enspace_{0.9262}^{0.9294}$  & $0.8339\enspace (\frac{83,392}{100,000})$  & $0.8946\enspace (\frac{44,732}{50,000})$  & $0.7732\enspace (\frac{38,660}{50,000})$  & $0.8801\enspace (\frac{38,660}{43,928})$  & $0.8232$ \\\cline{1-1}\cline{3-9} 
\multirow{3}{*}{FTS}                   &                          & $10^{-5}$ & $0.9200\enspace_{0.9182}^{0.9217}$             & $0.8215\enspace (\frac{82,153}{100,000})$   & $0.8888\enspace (\frac{44,440}{50,000})$       & $0.7543\enspace (\frac{37,713}{50,000})$     & $0.8715\enspace (\frac{37,713}{43,273})$             & $0.8087$    \\  \cline{3-9} 
                   &                          & $10^{-6}$ & $0.8669\enspace_{0.8646}^{0.8692}$             & $0.7752\enspace (\frac{77,519}{100,000})$   & $0.8165\enspace (\frac{40,823}{50,000})$       & $0.7339\enspace (\frac{36,696}{50,000})$     & $0.7999\enspace (\frac{36,696}{45,873})$             & $0.7655$    \\  \cline{3-9} 
                   &                          & $10^{-7}$ & $0.8126\enspace_{0.8099}^{0.8152}$             & $0.7247\enspace (\frac{72,468}{100,000})$   & $0.7502\enspace (\frac{37,509}{50,000})$       & $0.6992\enspace (\frac{34,959}{50,000})$     & $0.7368\enspace (\frac{34,959}{47,450})$             & $0.7175$    \\ \hline
\end{tabular}

    }
\end{table}

\end{document}